\PassOptionsToPackage{prologue,dvipsnames}{xcolor}

\documentclass[pdflatex,sn-mathphys-num]{sn-jnl}

\usepackage{graphicx}%
\usepackage{multirow}%
\usepackage{amsmath,amssymb,amsfonts}%
\usepackage{amsthm}%
\usepackage[title]{appendix}%
\usepackage{xcolor}%
\usepackage{textcomp}%
\usepackage{manyfoot}%
\usepackage{booktabs}%
\usepackage{adjustbox}
\usepackage{algorithm}%
\usepackage{lmodern}
\usepackage{algorithmicx}%
\usepackage{algpseudocode}%
\usepackage{listings}%
\usepackage{adjustbox}
\usepackage{booktabs}
\usepackage{xcolor}
\usepackage[dvipsnames]{xcolor}

\begin{document}

\title[Article Title]{Low-resource Information Extraction with the European Clinical Case Corpus}


\author*[1]{\fnm{Soumitra} \sur{Ghosh}}\email{sghosh@fbk.eu}
\author[3]{\fnm{Begona} \sur{Altuna}}\email{begona.altuna@ueu.eus}
\author[1]{\fnm{Saeed} \sur{Farzi}}\email{sfarzi@fbk.eu}
\author[1,2]{\fnm{Pietro} \sur{Ferrazzi}}\email{pietro.ferrazzi@phd.unipd.it}
\author[1]{\fnm{Alberto} \sur{Lavelli}}\email{lavelli@fbk.eu}
\author[1]{\fnm{Giulia} \sur{Mezzanotte}}\email{gmezzanotte@fbk.eu}
\author[1]{\fnm{Manuela} \sur{Speranza}}\email{manspera@fbk.eu}
\author[1]{\fnm{Bernardo} \sur{Magnini}}\email{magnini@fbk.eu}


\affil*[1]{Fondazione Bruno Kessler, Trento, Italy}

\affil[2]{University of Padua, Padua, Italy}

\affil[3]{GOI institute, Basque Summer University (UEU), Eibar, Spain}


\abstract{We present E3C-3.0, a multilingual dataset in the medical domain, comprising clinical cases annotated with diseases and test-result relations. The dataset includes both native texts in five languages (English, French, Italian, Spanish and Basque) and texts translated and projected from the English source into five target languages (Greek, Italian, Polish, Slovak, and Slovenian). A semi-automatic approach has been implemented, including automatic annotation projection based on Large Language Models (LLMs) and human revision. We present several experiments showing that current state-of-the-art LLMs can benefit from being fine-tuned on the E3C-3.0 dataset. We also show that transfer learning in different languages is very effective, mitigating the scarcity of data. Finally, we compare performance both on native data and on projected data. We release the data at \url{https://huggingface.co/collections/NLP-FBK/e3c-projected-676a7d6221608d60e4e9fd89}}




\maketitle
\section{Introduction}
\label{sec:introduction}

Information Extraction (IE) from text \cite{cowie1996information} aims at facilitating the analysis of the content of large document collections. However, data scarcity poses significant challenges for IE based on machine learning approaches, where labeled data (i.e., data where the correct annotations are provided) is crucial for several reasons.
First, data serves as the foundation for supervised learning, where algorithms learn to make predictions or to classify data based on input-output pairs. This process enables models to generalize from seen instances to unseen instances, improving predictions and decision-making. Additionally, labeled data is crucial for evaluating the performance of machine learning models, as it provides benchmarks to qualify the output of machine learning algorithms. 

In the paper, we address  data scarcity for IE in the medical field \cite{uzuner20112010}, with a focus on emergency departments (EDs). There are several reasons that motivate data scarcity in this domain: obtaining high-quality, annotated datasets is often difficult due to the large patient volume (emergency departments handle a large number of patients daily); time constraints (the urgent nature of emergency medicine means healthcare professionals have limited time to document patient information thoroughly); privacy concerns (patient data is highly sensitive
and its collection implies strict regulations such as GDPR,\footnote{General Data Protection Regulation in EU. \url{https://gdpr-info.eu/}} UK GDPR,\footnote{UK General Data Protection Regulation. \url{https://ico.org.uk/for-organisations/data-protection-and-the-eu/data-protection-and-the-eu-in-detail/the-uk-gdpr/}} and HIPAA\footnote{Health Insurance Portability and Accountability Act in the US.\url{https://www.hhs.gov/hipaa/index.html}}); data complexity (emergency department records often contain complex and unstructured data, including free-text notes, varied terminologies, and inconsistent formats), resource limitations (emergency departments typically
operate with limited resources and staff who are primarily focused on patient care rather than data
management and annotation), and the time-consuming process of manually labeling data.

In the paper, we present our experience to partially overcome the above issues. We approached data scarcity by developing a corpus of publicly available \textit{clinical cases}, which we consider a proxy for documents produced in EDs. We report on the design, creation, and experimentation of the European Clinical Case Corpus (E3C) \cite{magninietal21}, a freely available collection of annotated clinical cases in nine European languages.
E3C started with five languages as an initiative of the European Language Grid (ELG) project\footnote{The European Language Grid provides access to Language Technology resources from all over Europe. ELG contains tools and services, language resources and information on European LT companies and research organisations as well as their projects. \url{https://live.european-language-grid.eu}}, and was then extended to additional four low-resource languages (Greek, Polish, Slovak, and Slovenian) in the context of the eCREAM project (enabling Clinical Research in Emergency and Acute care Medicine)\footnote{eCREAM (\url{https://ecreamproject.eu}) is a European project coordinated by the Laboratory of Clinical Epidemiology at the Istituto di Ricerche Farmacologiche Mario Negri IRCCS (IRFMN). It involves 11 partners in 8 countries (France, Greece, Italy, Poland, Slovakia, Slovenia, Switzerland, and the United Kingdom).}.
A major objective of eCREAM is to develop state-of-the-art NLP technologies able to interpret Electronic Health Records (EHRs) content and extract crucial information (metadata) from them, which will then be used to make accurate analysis and prediction of EDs effectiveness. 


Information extraction for low-resource languages presents several research challenges. First, collecting data is typically a critical issue, due to their scarcity. However, the recent diffusion of Large Language Models (LLMs) has opened up new options for synthetic data generation \cite{li2023synthetic}. For extending the E3C corpus to new languages, we experimented the use of LLMs for both translation and annotation projection from a high resource language (i.e., English) to low-resource languages. We will show that, although in a context of high complex annotations (e.g., relations among clinical tests and their results), we were able to obtain high quality data through a semi-automatic process. Second, training IE models requires large amount of annotated data, and performance may critically depend on availability of enough representative data. Even in this case, we show that pre-trained LLMs, and particularly LLMs pre-trained in the medical domain, can be effective on two IE tasks (clinical entity detection and test-result relation extraction) even with a very limited number of training documents. In addition, we will show that cross-lingual transfer learning further improves performance on low-resource languages. Finally, we provide results and discussion of our comprehensive experimentation with LLMs on native data and translated data for IE tasks.


The paper is structured as follows. First, we report on the  two development phases of the E3C corpus. The first phase, (Section \ref{Sec:e3c-original}), includes  collection,  design and  manual annotation of clinical cases in five languages, i.e. English, Italian, French, Spanish, and Basque. The second phase (Section \ref{Sec:e3c-projected}) describes the semi-automatic extension of the E3C corpus to four additional languages (i.e., Greek, Polish, Slovakian and Slovenian). Section \ref{sec:information_extraction} introduces the two IE tasks, clinical entity detection and relation extraction. Then, there are three sections on IE experiments. Section \ref{sec:experiments_revised} reports experiments comparing performance on fully automatic synthetic data with data manually revised by experts. Section \ref{sec:experiment_all_languages} is about cross-language transfer, and, finally, Section~\ref{sec:italian_native_vs_projected} compares native data against synthetic data.

\section{The E3C Dataset: Native Languages}
\label{Sec:e3c-original}

The European Clinical Case Corpus (E3C) is a freely available\footnote{E3C is released under Creative Commons License Attribution 4.0 International (CC BY 4.0).} multilingual corpus encompassing five languages, i.e. English, Italian, French, Spanish, and Basque\footnote{The European Clinical Case Corpus (E3C) is available for download from the project's website: 
\url{https://e3c.fbk.eu/data} 
}. It consists of clinical narratives manually annotated with semantic information, thus allowing for linguistic analysis, benchmarking, and training of information extraction systems.

\subsection{Data Collection}


When building the E3C corpus, a major concern has been ensuring its reusability and shareability, which forced us to use anonymised and freely redistributable clinical cases. We collected journal abstracts available from PubMed\footnote{\url{https://pubmed.ncbi.nlm.nih.gov}} and extracted clinical cases published in journals (such as The Pan African Medical Journal) or available from medical training resources (e.g., the SPACCC: Spanish Clinical Case Corpus\footnote{\url{https://github.com/PlanTL-GOB-ES/SPACCC}}) and admission tests for specialties in medicine.

\begin{figure}
    \centering
    \includegraphics[scale=0.4]{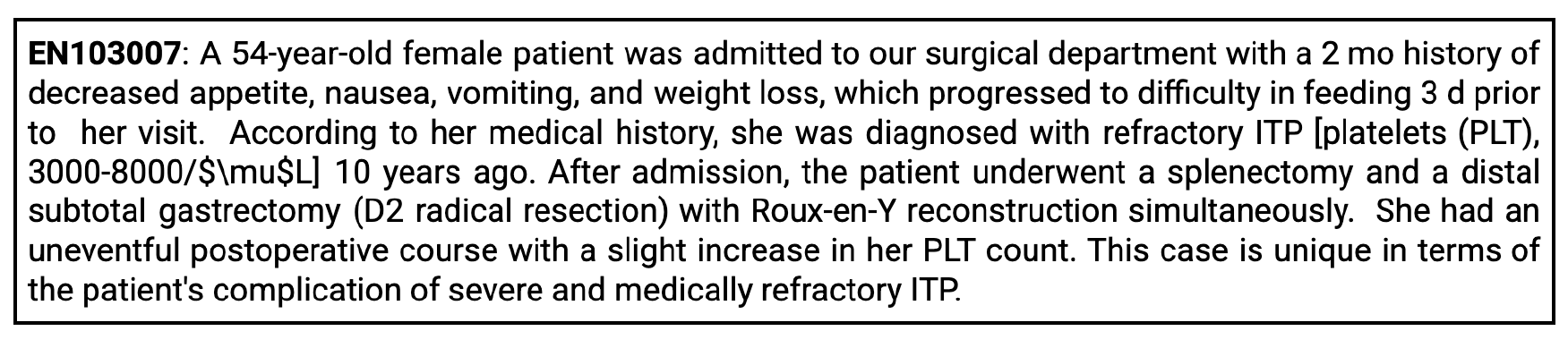}

    \caption{A clinical case example from the English E3C dataset (document EN103007).}

    \label{fig:c_n}
\end{figure}

A clinical case is a statement of a clinical practice focusing on a single patient; E3C focuses on this specific type of clinical narrative because they are rich in clinical entities as well as temporal information, which is almost absent in other clinical documents (e.g., radiology reports). As shown in Figure~\ref{fig:c_n}, the E3C's clinical cases typically start presenting the reason for a clinical visit (i.e., the patient's symptoms) and then describe the assessment of the patient’s situation; physical exams and laboratory tests play a central role in diagnosing diseases and disorders, therefore they are reported in clinical cases and their results meticulously documented. The final diagnosis usually conclude the text, but treatment, outcome, and follow-up may be present as well.

Symptoms, tests, observations, treatments, and diseases are all marked in E3C as they are relevant events for the history of a patient, and to understand the evolution of a patient's health it is relevant to place them in chronological order, so temporal relations are also made explicit. 
Since precision in symptom description and diagnosis is utterly important in the clinical field, clinical findings, body parts, laboratory results, and measurements, etc., are identified as well.




\subsection{Semantic Annotations} \label{e3c-annotations}

E3C foresees two types of semantic annotations: 
\begin{enumerate}
\item temporal information, i.e., events, time expressions, and relations between them;
\item clinical entities (such as pathologies and symptoms), body parts, laboratory results (in relation to the test event they refer to), and actors.
\end{enumerate}

In both cases, we have different categories of span-based annotations and different types of relations (relations link two span-based annotations that could be of the same or different category). 



\begin{figure}[!ht]
    \centering
    \includegraphics[width=1\linewidth]{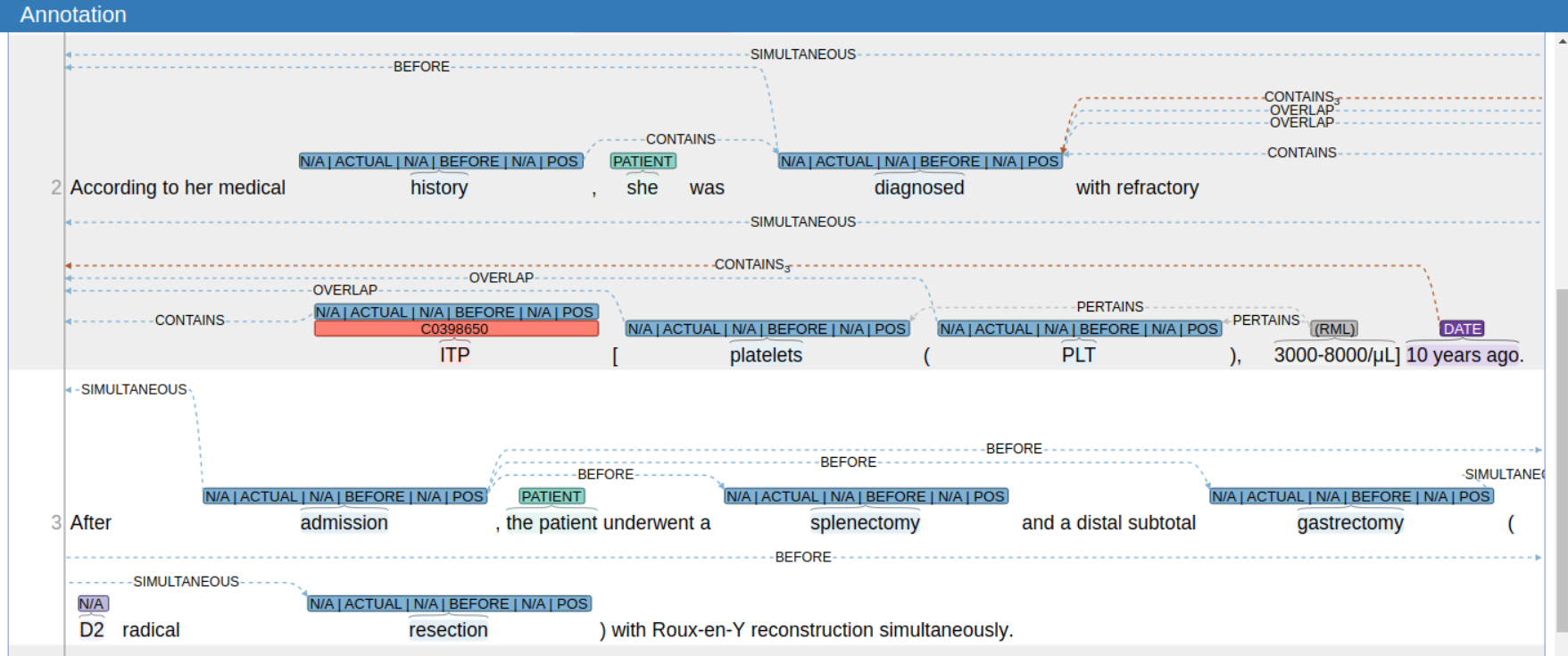}
    \caption{A sample E3C annotated text (excerpt from EN103007.xml) is displayed using the WebAnno annotation tool. Annotations are highlighted: clinical entities (red), results/measurements (grey), events (blue), temporal expressions (purple), actors (turquoise), and body parts (lilac).}
    \label{fig:e3cAnnotations}
\end{figure}

The E3C annotation framework is extremely rich in terms of attributes assigned to the different annotation categories and in terms of relation types (as shown in Figure \ref{fig:e3cAnnotations}).
For the sake of simplicity, in Table \ref{tab:categories} we report the annotation categories (with descriptions and examples) without specific attributes.


\begin{table}[!ht]
\centering
\caption{E3C annotations based on textual spans.}\label{tab:categories}
\renewcommand{\arraystretch}{1.1}
\begin{tabular}{l|p{3cm}|p{6cm}}
\hline
\textbf{Category} & \textbf{Description} & \textbf{Example} \\ \hline
CLINICAL ENTITY      & disorders, pathologies, and symptoms &  ``ITP" (an acronym for Immune thrombocytopenia), ``dicreased appetite", ``nausea"\\ \hline
BODYPART      & parts of the human body & ``D2" (the descending part of duodenum)\\ \hline
 RML           & (often numeric) results and measurements & ``3000-8000/$\mu$L" \\ \hline
 ACTOR      & any person or animal mentioned in the text &   ``A 54-year-old female patient", ``she", ``the patient" \\\hline
 EVENT     & events & ``history", ``diagnosed", ``ITP", ``splenectomy" \\ \hline
 TIMEX3     & time expressions & ``10 years ago" \\
      \hline
\end{tabular}
\end{table}

\begin{table}[!ht]
\caption{E3C relations (high-level classification).}\label{tab:relations}
\centering
\renewcommand{\arraystretch}{1.1}
\begin{tabular}{l|l|p{2.5cm}|p{3.7cm}}
\hline
\textbf{Relation} & \textbf{Source} & \textbf{Target} & \textbf{Examples} \\ \hline
PERTAINS-TO   & RML & laboratory test or measurement event & ``3000-8000/$\mu$L" pertains-to ``platelets" \\ \hline
TLINK (temporal link)  & event/TIMEX3 & event/TIMEX3 & ``admission" before ``splenectomy", ``10 days ago" contains ``diagnosed"  \\ \hline
ALINK (aspectual link) & event & event & ``started" initiates ``diet" \\  \hline
\end{tabular}
\end{table}

As far as the classification of relations is concerned (see Table \ref{tab:relations}), E3C foresees three different types:
\begin{itemize}
\item PERTAINS-TO, i.e. relations linking laboratory and test results and measurements (RMLs) to the span-based annotation of the laboratory tests and measurements (events) from which they were obtained;
    \item TLINKs, i.e. temporal expressions holding between events and/or time expressions (by expressing precedence, overlap, containment, initiation or ending between two events and/or time expressions, TLINKs allow for chronologically ordering them); 
\item ALINKs, i.e. relations that link aspectual events, i.e. events indicating a specific phase (beginning, end, continuation, etc.) of an event, to the event itself;
\end{itemize}


The E3C corpus was exploited as a benchmark for training and assessing entity recognition models~\cite{zanoli2023assessment}.
In these experiments, traditional machine learning models like Conditional Random Fields (CRFs) and more recent multilingual pre-trained models based on deep learning were compared with standard baselines.


More recently, two shared tasks took advantage of the E3C data, namely CLinkaRT\footnote{\url{https://e3c.fbk.eu/clinkart}} \citep{altuna2023clinkart}  and TESTLINK\footnote{\url{https://e3c.fbk.eu/testlinkiberlef}} \citep{altuna2023overview}, organised in the EVALITA 2023 and IberLEF 2023 events, respectively. CLinkaRT (for Italian) and TESTLINK (for Spanish and Basque) focused on evaluating different systems on the task of identifying the RMLs and the laboratory tests or measurements they refer to (PERTAINS-TO relations).

\section{Semi-automatic Extension of E3C to New Languages}
\label{Sec:e3c-projected}



The second phase of the development of E3C consists of a semi-automatic extension. By using the English dataset as a source, 
we automatically translated the texts and projected the manual annotations, to create new annotated datasets in five different languages: Greek, Polish, Slovak, Slovenian and Italian.

\subsection{Translation and Annotation Porting}

The idea of reducing the effort needed to produce new datasets in new languages by transferring semantic annotations from one language to another is already present in the literature, where annotation projection has often been formulated as the task of transporting, on parallel corpora, the labels pertaining to a given span in the source language into its corresponding span in the target language; it is basically a task consisting of three steps, translation, alignment, and annotation transfer \cite{bentivoglietal04, garcia-etal2023}. 




The task is quite challenging 
because concepts are expressed in different languages with linguistic constructions that might differ at various linguistic levels, ranging from syntax to morphology.
From the syntactic point of view, for instance, we can have languages (such as English) where the subject of a verb is always expressed and languages (such as Italian, for example) where a pronominal subject can be omitted; in terms of transferring E3C annotations, this is an interesting problem as the pronoun in question (present in the English source text) might refer to the patient, therefore, be marked as an ACTOR; in the impossibility of finding the pronoun, the automatic system cannot possibly transfer the annotation.
Possible problems occurring at the morphological level are represented by lexical gaps,  i.e. words that do not have a direct equivalent in the target language 
(for instance, translating ``successfully" into Italian requires the combination of words corresponding to ``with success"),
as, translating an adverb using combinations of words belonging to different grammatical categories, might impact the cross-lingual annotation transfer.

\begin{figure}[!ht]
    \centering
    \includegraphics[width=1\linewidth]{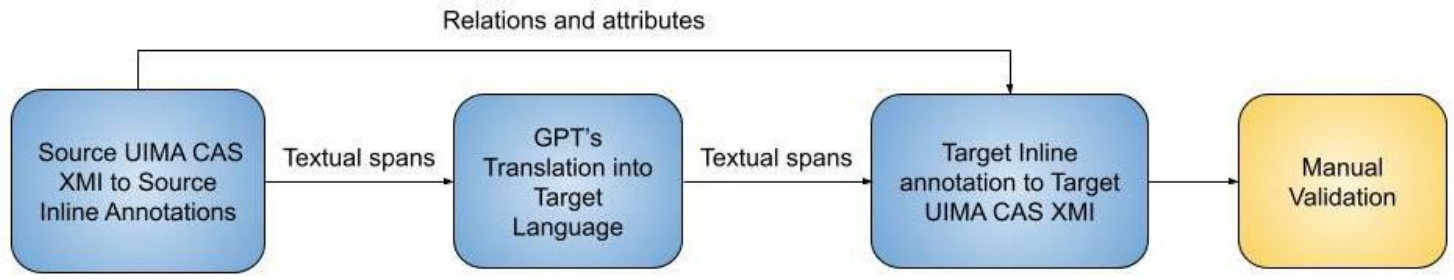}
    \caption{Procedure for the extension of an E3C annotated dataset to a target language.}
    \label{fig:phases}
\end{figure}

 The projection procedure implemented for the extension of E3C consists of two main phases:

  \begin{enumerate}
     \item an automatic procedure translates the texts from English to a new language while simultaneously transferring the annotations as well (the three main steps of this procedure are illustrated in Figure~\ref{fig:phases});
     \item the erroneous annotations and those that the system has not been able to transfer are corrected/added manually with the help of a native speaker.
 \end{enumerate}

\begin{figure}[!ht]
    \centering
    \includegraphics[width=1\linewidth]{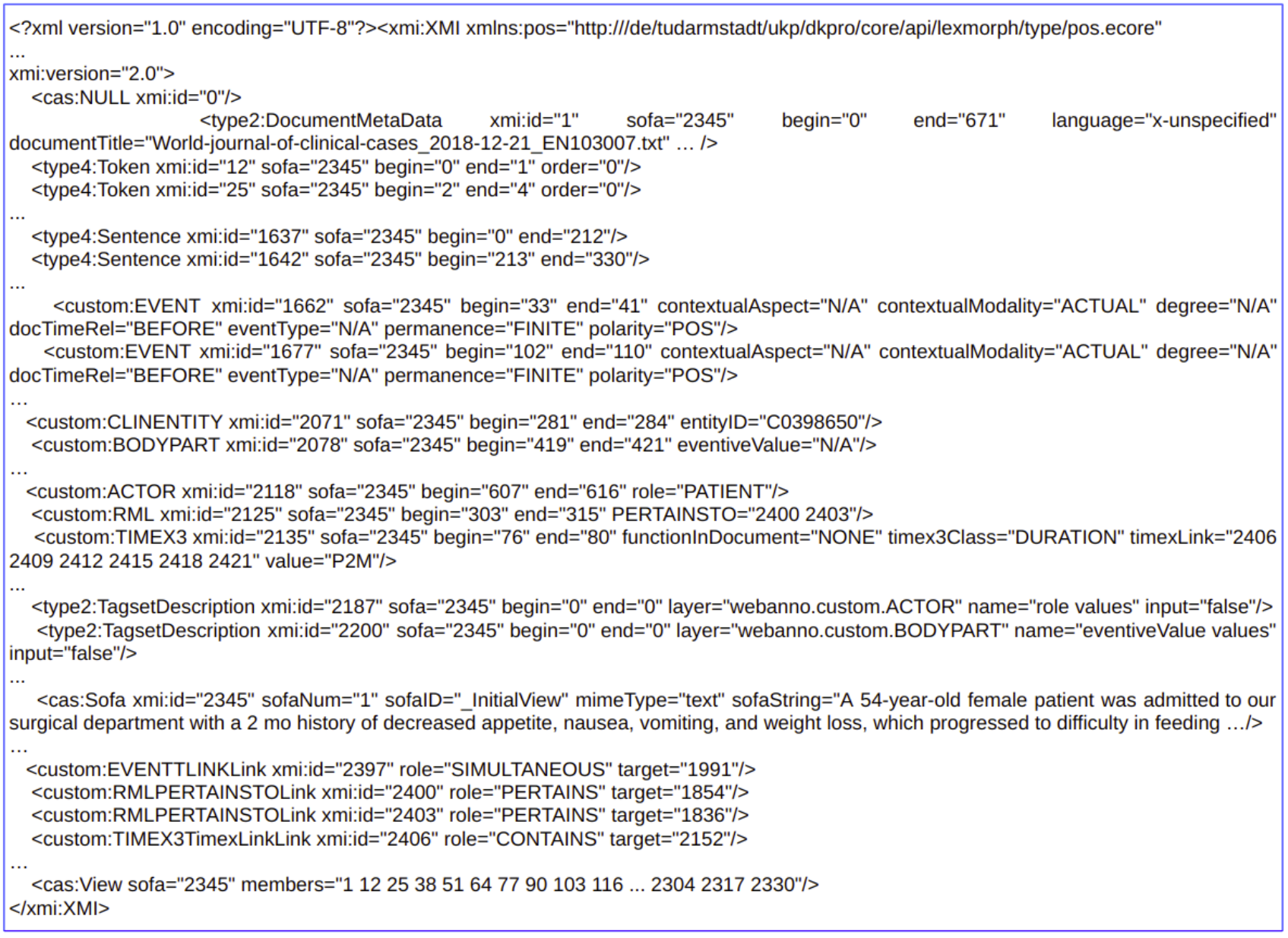}    
    \caption{Example of the stand-off XMI for clinical case EN103007.xml in Figure~\ref{fig:il_en}. For the sake of space and simplicity of understanding, the figure presents a reduced content of the actual XMI.}
    \label{fig:standoff}
\end{figure}

\subsection{Automatic projection procedure}
\label{automaticprocedure}

\begin{figure}
    \centering
    \includegraphics[scale=0.38]{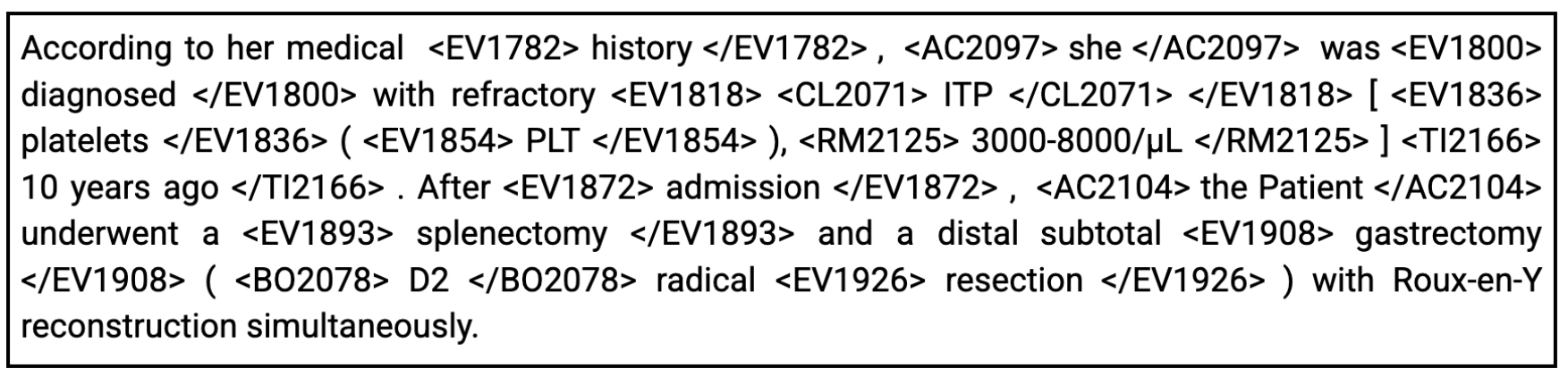}

    \caption{Inline representation of the stand-off XMI from Figure~\ref{fig:standoff}.}

    \label{fig:il_en}
\end{figure}

To proceed with the annotation transfer task, we generate inline annotations from E3C standard format (as shown in Figure \ref{fig:phases}). E3C is originally distributed in the UIMA CAS XMI\footnote{The UIMA CAS XMI format (Common Analysis Structure eXtensible Markup Language) is a standard data serialization format used within the Apache UIMA (Unstructured Information Management Architecture) framework. It enables the representation and exchange of structured data annotations over unstructured text or other forms of unstructured data.\url{https://uima.apache.org/}} standoff format (shown in Figure~\ref{fig:standoff}), where annotations are stored separately from the text and associated with the text using begin and end offsets. This format allows for a straightforward machine representation of annotation spans, as well as their attributes and relations, using any annotation tool (WebAnno\footnote{\url{https://webanno.github.io/webanno/}} in our case). Our goal is to produce an equivalent and parallel target language XMI from the source XMI by leveraging powerful LLMs (such as, for example, OpenAI's GPT models). However, this is not straightforward due to the complex structure of XMI (as evident from Figure~\ref{fig:standoff}), which contains a variety of information beyond annotations, making direct porting very difficult even for state-of-the-art LLMs. To address this, we transform the source XMI into a simpler inline annotation format (shown in Figure~\ref{fig:il_en}), filtering out attributes and relations and storing only span-based annotations within the annotated text using opening and closing tags. The main challenge is the complexity of the E3C annotation schema, which allows for overlapping and nested entities. We achieve this conversion using a stand-alone Python script that transfers the span-based annotations, maintaining all nested entities without losing annotations.


\begin{figure}[!ht]
    \centering
    \includegraphics[scale=0.43]{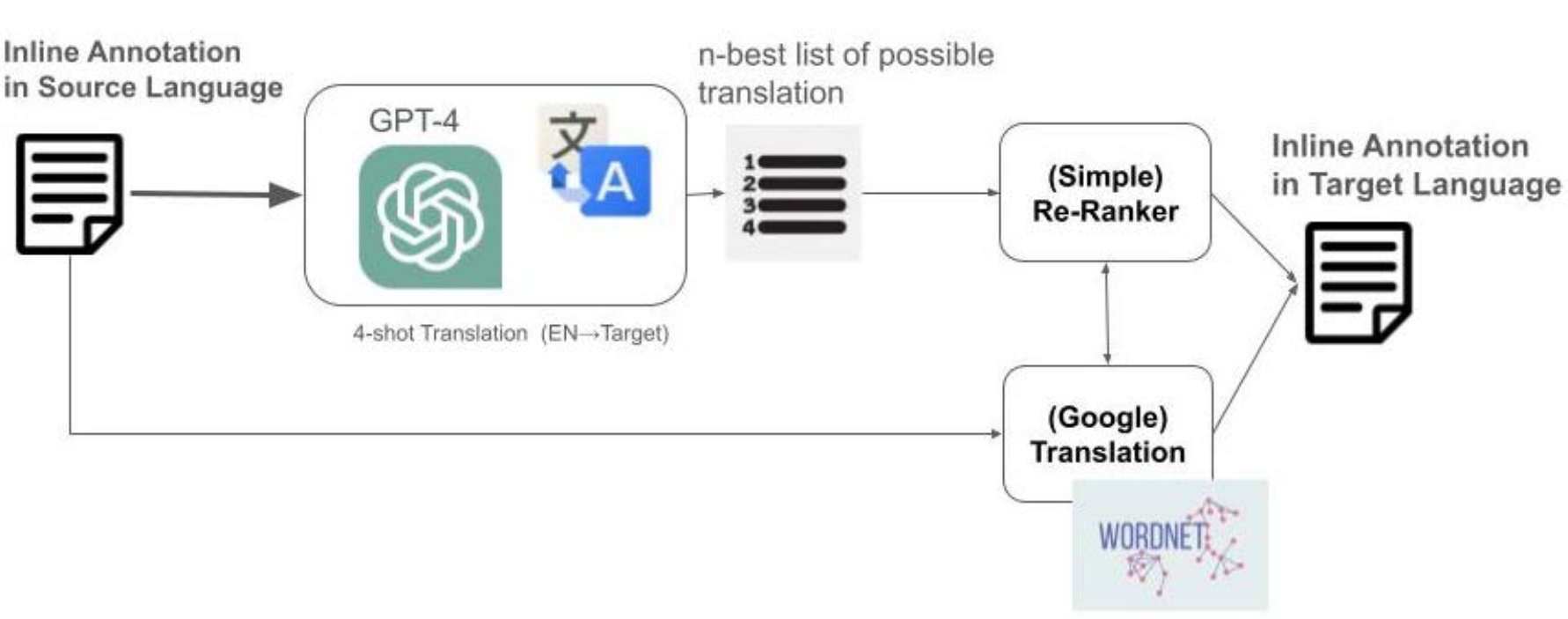}
    \caption{Translation system architecture. }
    \label{fig:tf}
\end{figure}

\begin{figure}
    \centering
    \includegraphics[scale=0.43]{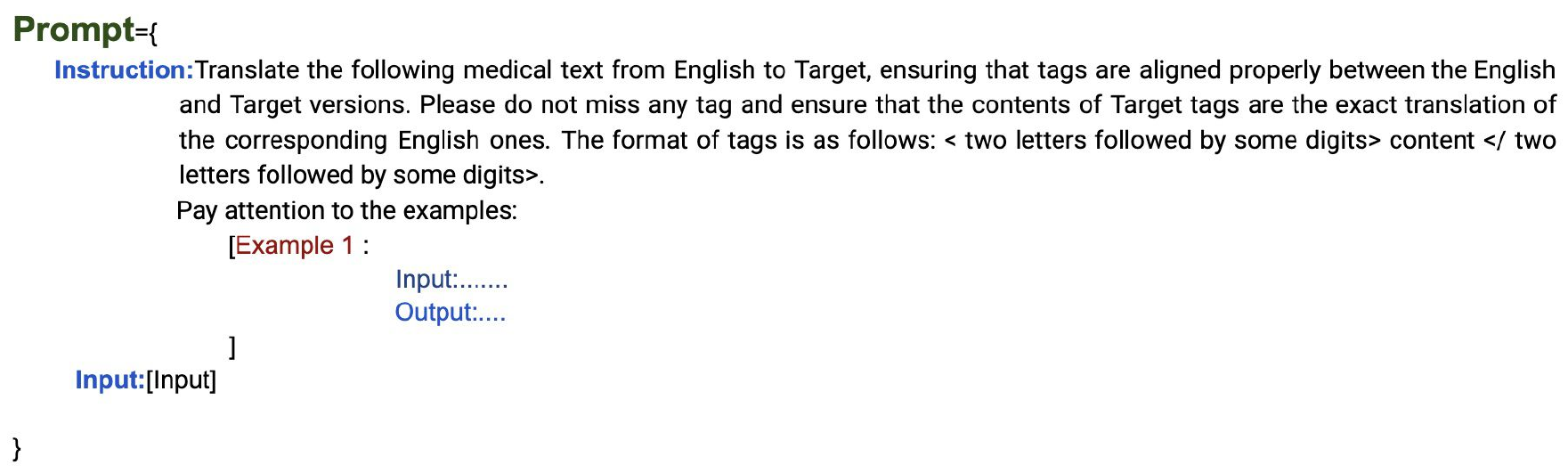}
    \caption{The few-shot prompt used to translate English notes to target language.}
    \label{fig:pt}
\end{figure}

Inline annotated data is suitable for feeding a translation system based on GPT-4\footnote{gpt-4-0125 from Open AI platform has been employed to translate texts.}, whose overall architecture is shown in Figure~\ref{fig:tf}. Such system has been developed to translate the clinical cases (English texts) into target languages (i.e. Italian, Slovak, Slovenian, Polish, and Greek), while simultaneously transferring the annotations to the translated texts.
As shown in Figure~\ref{fig:tf}, to increase the translation quality and input GPT-4 with guidelines on how to transfer annotations, a few-shot learning approach~\citep{mann2020language} has been employed. The prompt for few-shot annotation projection learning is depicted in Figure~\ref{fig:pt}.

The translation system shown in Figure~\ref{fig:tf} leverages GPT-4 to generate accurate translations of an English medical case containing specific tags into a target language. The system employs a 4-shot prompting strategy to guide GPT-4 in understanding the context and structure of the translation task, ensuring higher-quality outputs. GPT-4 produces an n-best list of four possible translations to provide multiple options. A simple re-ranking mechanism evaluates these translations based on the number of mismatched or missing tags. This evaluation is supported by Google Translate\footnote{\url{https://translate.google.com/}} and WordNet\footnote{WordNet is a comprehensive lexical database of the English language developed at Princeton University, which organizes words into sets of synonyms called synsets. Each synset represents a single concept and includes definitions and usage examples \cite{miller1995wordnet}.}, which help detect and quantify discrepancies in medical tags. The 4-shot prompt and the generation of 4-best translations strike a balance between computational cost\footnote{On average, translating a medical case cost us one Euro, while the price of GPT-4-0613 for input tokens was $0.03$ per 1,000 tokens and for output tokens was $0.06$ per 1,000 tokens at the time of dataset translation in August 2024.} and translation effectiveness, ensuring robust and accurate handling of complex medical tags while keeping it affordable and efficient. The system's ultimate goal is to produce a final translation that minimizes errors, ensuring that all medical tags are correctly translated and aligned with the original English text. The prompt used consists of three parts: an instruction, an input, and four examples. The instruction section explains translation tasks along with introducing the tags' format. Selecting examples for few-shot learning is crucial \cite{wang2020generalizing} because the model's ability to generalize effectively from limited data hinges on the representativeness and diversity of the chosen examples. High-quality examples that are consisting of core patterns of the translation task enable the model to form meaningful representations, improving its capacity to handle unseen sentences and tags. Conversely, poorly selected examples can lead to biased learning, overfitting, or suboptimal performance \cite{xu2022alleviating}. In this context, we focus on choosing samples that consist of a diverse range of tags, i.e EVENT, BODY, RMLS and so on  and tag types, including nested  and overlapping tags.

\begin{figure}
    \centering
    \includegraphics[scale=0.4]{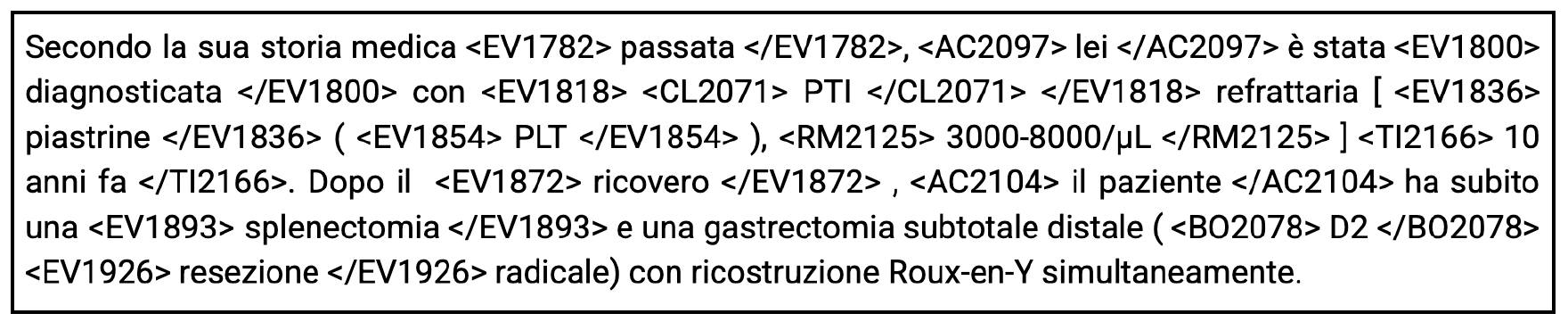}
    \caption{The Italian translation of the text in Figure \ref{fig:il_en}.}
    \label{fig:il_t}
\end{figure}

The resultant translations are the inline annotations in the target language that need to be converted back 
to the original XMI format (see for example the Italian output in Figure~\ref{fig:il_t}).
To achieve this, we construct the target XMI by first extracting language-independent information from the source XMI (such as rows that define the file's structure, preamble and closing tags, metadata, etc.). Next, we generate the annotation rows in the target XMI, 
exploiting the annotation tool WebAnno for tokenization and
calculating appropriate offsets for each annotation span based on the target language inline annotations. For each created annotation row in the XMI, we also populate attributes associated with that annotation, retrieving them from the source XMI (as this information was intentionally removed when preparing prompts for translation). Finally, we copy all relations from the source XMI to the newly created target XMI files. Coherency between annotations and relations in the source and target XMIs is maintained using unique XMI identifiers. To facilitate manual validation (discussed in detail in the following section), any missing or mismatched annotation information is intuitively presented in the target XMI through additional metadata rows (for missing 
annotations)
and 
flagged by means of specifically created
attributes (for mismatches).

\subsection{Translation and projection quality measures}
\label{MTQ-measures}
To assess the quality of the output of the procedure described in Section~\ref{automaticprocedure} 
a comprehensive evaluation framework is employed. 

\paragraph{Translation quality}
Translation quality is calculated using BERTScore \cite{zhang2019bertscore}.
To calculate BERTScore, reference sentences in the target language are typically required. However, since such references are not available, back-translation is used to translate the target text back into the source language, enabling a comparison between the original and the back-translated source texts. This approach is not ideal because it introduces compounding translation errors: errors from the initial source-to-target translation and additional errors from the target-to-source back-translation. These errors can distort the evaluation of semantic and structural accuracy. However, despite its limitations, this method provides a practical way to approximate the quality of the translation. By comparing the original source text with its back-translated counterpart, we gain a valuable insight into how effectively the system preserves meaning and structure during translation. To this end, our assessment considers `missing tags' and `mismatched tags' ratios to assess tag preservation  in the translated text.

\emph{\textbf{Missing tags:}} A \textit{missing tag} is a source tag that does not appear in the translated text. 

\emph{\textbf{Mismatched tags:}} A \textit{mismatched tag} is a source tag that appears in the target text but with an incorrect span of text. 

For instance, considering the text shown in Figure~\ref{fig:il_en} as input and the text shown in Figure~\ref{fig:il_t} as output, the tag \textless EV1908\textgreater \textless/EV1908\textgreater is missing tag while \textless EV1782\textgreater...\textless/EV1782\textgreater is mismatched tag. 

While identifying missing tags is straightforward, detecting mismatched tags involves translating the span back into the source language and using a lexical database like WordNet to account for synonymous phrases. To this end, Google Translate\footnote{\url{https://translate.google.com/}} is employed for span translation, while English WordNet ensures accurate recognition of synonyms, enabling a robust evaluation of both semantic accuracy in machine translation systems. 

\begin{figure}[!ht]
    \centering
    \includegraphics[scale=0.44]{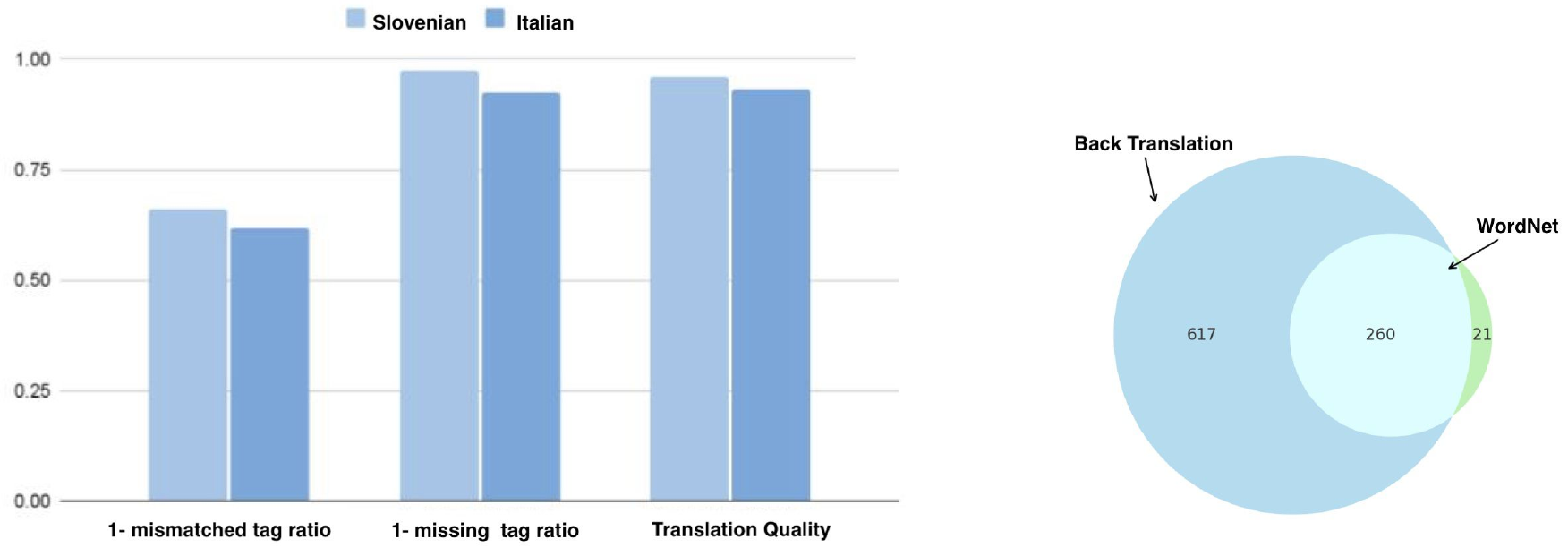}
    \caption{Left: Translation quality and the number of mssing and mismatched tags for Italian and Slovenian (10 docs). Right: A comparison of back-translation and WordNet-based approaches as the number of detected mismatched tags.}
    \label{fig:figure_table}
\end{figure}

As shown in Figure~\ref{fig:figure_table} (Left), both Slovenian and Italian show strong performance in terms of translation quality and the number of missing and mismatched tags. Slovenian showing a slight edge in the number of missing and mismatched tags, and Italian excelling marginally in overall translation quality. Although the metrics may vary across languages due to differences in language complexity and the translation quality of GPT-4, analysis of various languages shows that performance remains generally consistent and within an acceptable range for human annotation. According to the project pipeline, the handling of missing and mismatched tags is deferred to 
native speakers
during the 
manual revision
phase.

Figure~\ref{fig:figure_table} (Right) demonstrates the contribution of Wordnet in boosting the performance of the proposed system.  
In the proposed translation and porting system, GPT-4 is tasked with generating a 4-best translation list for a given input. This strategy aims to improve tag fidelity by employing a re-ranking algorithm to select the sentence with the lowest scores for missing and mismatched tag ratios among the four candidates. Since the translation quality across the 4-best outputs is generally consistent, and back-translation is computationally intensive, translation quality metrics are not used in the re-ranking process. Additionally, the decision to limit GPT-4 to producing only four outputs is driven by the cost of GPT-4 usage.

Table \ref{tab:porting_errors_stat} presents detailed statistics on candidate mismatches and missing tags produced by the projection procedure across five languages: Slovak, Italian, Slovenian, Greek, and Polish. The table categorizes errors into four types—CL, EV, RML, and Other (Oth.)—with a total (TOT) count for each language.

Among the mismatches, Polish exhibits the highest total count (3524), followed by Greek (3299) and Italian (3032), while Slovenian records the lowest (2727). The EV category consistently accounts for the largest share of mismatches across all languages, with Polish (1750) and Greek (1727) showing the highest values, whereas Slovenian has the lowest (1171). The RML category, which represents a smaller proportion of mismatches, varies across languages, with Slovenian (110) showing the highest count and Slovak (39) the lowest. The Other (Oth.) category also contributes significantly to mismatches, particularly in Polish (1193) and Slovak (1123).

For missing tags, Italian has the highest total count (851), followed by Slovenian (338) and Greek (345), while Polish records the lowest (265). The EV category again shows the highest number of missing tags across all languages, with Italian (750) leading, followed by Greek (304) and Slovenian (297). In contrast, the RML category exhibits minimal occurrences, with Polish and Slovak each reporting only one and two missing tags, respectively. The Other (Oth.) category remains relatively small but still contributes to missing errors, with Italian (47) and Polish (20) showing notable counts. 
These variations in error distributions suggest language-dependent challenges in maintaining tags during translation.

\begin{table}[!ht]  
\centering
\caption{Statistics about the candidate errors produced by the projection procedure. \textit{Abbreviations:} CL-Clinical Entities, EV-Events, RML-Test results and measurements, Oth.-included Body Parts, Actors and Timex.}
\label{tab:porting_errors_stat}
\begin{tabular}{l||r|r|r|r|r||r|r|r|r|r}
\hline
   & \multicolumn{5}{c||}{Candidate mismatches} & \multicolumn{5}{c}{Missing} \\
          &    CL &   EV  & RML  & Oth.  &  TOT & CL &  EV & RML & Oth. & TOT \\  \hline
Slovak    &   340 & 1397  &  39  & 1123  & 2899 &  8 & 227 &  2  & 13   & 250 \\ 
Italian   &   386 & 1660  &  48  &  938  & 3032 & 41 & 750 & 13  & 47   & 851 \\  
Sloven.   &   332 & 1171  & 110  & 1114  & 2727 & 13 & 297 &  7  & 21   & 338 \\
Greek     &   524 & 1727  &  84  &  964  & 3299 & 13 & 304 & 12  & 16   & 345 \\
Polish    &   518 & 1750  &  63  & 1193  & 3524 &  7 & 237 &  1  & 20   & 265 \\
\hline
\end{tabular}
\end{table}


\subsection{Manual validation}

The output produced by the (fully automatic) projection procedure needs some manual intervention as it contains a number of erroneous annotations, in addition to missing annotations that need to be added; the person needed for this task has to be a native speaker of the target language proficient in English as well.


The 
effort required, 
however,
is highly reduced by the fact that the task does not require revising the whole annotated dataset in the target language, but only the annotations that have been previously 
flagged
as mismatches, i.e. annotations projected with a wrong textual span
(as defined in Section~\ref{MTQ-measures}).
With regard to these annotations, 
annotators 
are provided with not only the original English textual span but also all the information about the English source annotations. This means that, while familiarity with the E3C annotation guidelines is still desirable, the validation task is strongly simplified as, in the case of errors to be fixed, annotation attributes can be transferred mechanically from the English native version. 


In the end, the validation of each annotation flagged as mismatch simply consists of verifying the correctness of the textual span provided by the system by checking if the annotated portion of text in the translated document corresponds to the original English span; if not, annotators mark the correct text and add the attributes and relations where appropriate. 


Similarly,
in the case of missing annotations to be added,
annotators 
refer to
the span of the 
source
English annotation and to the E3C guidelines to correctly identify the span and on the information about the English source annotations, i.e. its attributes and the relations in which it is involved, to complete the annotation.



With respect to the rich set of E3C annotations described in Section~\ref{e3c-annotations}, this study focuses on two subtasks (the detection of clinical entities and the extraction of Pertains-to relations, as explained in detail in Section~\ref{tasks}), so the manual revision focused on Clinical entities, RMLs and a subset of events, i.e. those involved in a Pertains-to relation.
Task completion required a total of two working days per language.

As shown in Table~\ref{tab:porting_revision_stat}, the Greek and Polish datasets had the highest number of mismatches to be checked (above 700), but the effort was still limited. In most cases, in fact, the candidate mismatches were actually correct and annotators were just required to validate them.
The error rate is particularly low for Greek (and Slovak as well), where less than 15\% of the candidate mismatches actually had to be corrected.
Italian and Slovenian had the highest error rate but also a smaller number of candidate mismatches, so the final number of human interventions was still less than 300.

\begin{table}[!ht]
\caption{Statistics about revision of projected annotation of the E3C corpus.}
\label{tab:porting_revision_stat}
\begin{tabular}{l|r|r|r|r|r} 
\hline
                      & \textbf{Slovak} &  \textbf{Italian} & \textbf{Slovenian} & \textbf{Greek}  & \textbf{Polish} \\ \hline
\textit{Total mismatches}      & 2899   & 3032     & 2727      & 3299   & 3524 \\
\textit{Checked mismatches}    & 483    & 615      & 573       & 768    & 742 \\
\textit{Corrected mismatches}  & 71     & 274      & 236       & 102    & 207 \\
\textit{Error rate on checked} & 14.6\% & 44.5\%   & 41.2\%    & 13.2\% & 27.8\% \\ \hline
\textit{Total missing}         & 250    & 851      & 338       & 345    & 265 \\
\textit{Created missing}       & 23     & 148      &   47-152        & 64     & 29 \\ \hline
\end{tabular}

\end{table}


Candidate mismatches can be grouped into two main categories; on the one hand, we have what we call linguistic errors, where the projection task was made tougher by (sometimes also domain-specific) linguistic issues, and, on the other hand, we have procedural errors, i.e. cases where GPT-4 produced some errors in spite of linear translations between the two languages, possibly because of the excessive nesting of labels. In the following, we will focus on linguistic errors.

GPT-4 was able to correctly transfer the textual span even in cases where the translation was not completely straightforward (and therefore selected for manual revision). For instance, the word ``headache" (annotated as a clinical entity) was (correctly) translated by means of a three-word expression 
(``mal di testa", which has a different syntactic structure) and the expression ``60mm x 50mm across" (marked as an RML) was translated as ``60mmx50mm" (where the omission of the word ``across" is more than acceptable);
despite the complexity, the model had no problems in transferring both annotations. 
The same holds for even more complicated cases where we have so-called free translations.
A syntactically simple structure like ``generalised bodyache" (annotated as a clinical entity) has been translated as ``dolori diffusi in tutto il corpo" (which literally corresponds to ``pains widespread in the whole body") and still the annotation transfer worked out well.

In other cases, inevitably, the automatically produced output actually required some manual intervention. For instance, in ``calcified masses" (translated as ``masse calcificate"), we had two distinct clinical entities (``calcified" and ``masses") but the Italian output erroneously resulted in one single entity whose textual span covered the whole expression; this is probably due to the different word order in the two languages as in Italian, unlike in English, an adjective generally comes after the noun it refers to (and in fact ``masse calcificate" literally corresponds to ``masses calcified").
The impact of this linguistic phenomenon is even more evident with more complex (yet quite frequent) syntactic structures encompassing entities with overlapping spans. Let's take, as an example, the expression ``anterior and posterior capsular rupture",
where we have two clinical entities: 
E1 ``posterior capsular rupture" (whose textual span is straightforward) and 
E2 ``anterior [...] capsular rupture" 
(whose extent ``anterior and posterior capsular rupture" also includes the two extra words ``and posterior")
\footnote{According to the E3C guidelines, it is not admitted to mark discontinuous text spans like ``\underline{anterior} and posterior \underline{capsular rupture}"; if it is the case, all the words between those interested by the annotation must also be included, and the annotation is marked as ``discontinuous".}. 
In the Italian output E1  was missing, so we had to add the corresponding annotation (``rottura capsulare anteriore e posteriore")
\footnote{Notice that it contains the extra words ``anteriore e" and it is marked as ``discontinuous".},
while for E2, which was transferred as ``rottura capsulare anteriore e posteriore", we had to reduce the textual span to obtain `rottura capsulare anteriore".











\section{Information Extraction on the E3C Dataset}
\label{sec:information_extraction}

This section introduces the two IE tasks, clinical entity detection and relation extraction, on which we performed our experiments.
In particular, we describe the E3C dataset, the E3C tasks, the specific language models used in the experiments, and the fine-tuning procedure.

\subsection{Large Language Models for Clinical Information Extraction}
\label{llms}
Large Language Models (LLMs) have recently emerged as state-of-the-art models in a variety of tasks in the biomedical domain. Notable performances have been recorded in question answering \citep{singhal2023large, chen2023meditron}, document classification \citep{luo2022biogpt, lewis-etal-2020-pretrained}, sequence labeling \citep{bioNLP2024}, relation extraction \citep{gururangan-etal-2020-dont}, feature extraction \citep{huang2020clinicalbert} and other tasks. In the context of this work we focus on encoder-decoder and decoder-only models, as they are suitable for a wider range of generative tasks compared to encoder-only models \cite{10.1007/978-3-030-90072-4_23} and of the observed trend of decoder-only model reaching state-of-the-art results.
 

Examples of encoder-decoder models include T5 \cite{t5paper}, its multilingual version mT5 \cite{xue-etal-2021-mt5}, and BART \cite{lewis-etal-2020-bart}, which re-frame all tasks into a unified text-to-text format, making them particularly powerful when requiring precise input-output mappings. 

Decoder-only models have demonstrated remarkable capabilities in generating coherent and contextually relevant output, proving their strengths when demanding structured output after additional tuning. Such models are often divided into two categories based on the availability of the weights, namely close-source (GPT \cite{DBLP:journals/corr/abs-2303-08774}, Claude \cite{TheC3}, Gemini \cite{geminiteam2024geminifamilyhighlycapable}) and open-source (Llama \cite{DBLP:journals/corr/abs-2407-21783}, Mistral \cite{jiang2023mistral7b}, Qwen \cite{yang2024qwen2technicalreport} etc.), the latter offering transparency and greater control, enabling tailored usage for specific needs.


\paragraph{Large Language Models for Information Extraction}

LLMs have emerged as powerful tools for Information Extraction (IE) \cite{Xu2024}, thanks to their powerful architectures and training procedures that make them able to derive structured knowledge from unstructured natural language text. Leveraging their exceptional capabilities in understanding and generating text, LLMs have been integrated into various IE tasks, often adopting a generative paradigm. 
These tasks include sequence labeling \cite{cohan-etal-2019-pretrained}, entity detection \cite{10.1093/bioinformatics/btae163}, and relation extraction \cite{detroja2023survey}, where LLMs have shown remarkable potential to streamline processes and improve accuracy \cite{Xu2024}. Recent advancements have categorized these efforts into distinct subtasks, highlighting the innovative techniques employed to harness LLMs for IE.

\subsection{Data}
\label{data}
The E3C dataset has been developed during a four-year period leveraging different methodologies, reported in Sections \ref{Sec:e3c-original} (data on native languages), and \ref{Sec:e3c-projected} (data on projected languages). Table~\ref{tab:e3c_datasets} summarizes the E3C composition as for October 2024, in terms of number of documents (each document is a clinical case) that are annotated with both clinical entities and test-result relations. \textit{Native} refers to data originally provided in a language, while \textit{projected} refers to data that have been semi-automatically translated and projected using English as a source. This means that all the projected data is aligned (document and sentence level) with the English native, and that a consistent train-validation-test split among different languages is maintained. On the other hand, the native data, while being comparable (i.e., they are all clinical cases), are not aligned among languages. Notice that for Italian E3C includes both a native and a projected dataset, a situation that we will analyze in Section \ref{sec:italian_native_vs_projected}.


\begin{table*}[!ht]
 \centering
\caption{E3C-3.0 available datasets, with number of documents and total number of tokens, for both native and projected data.}
\label{tab:e3c_datasets}
\begin{adjustbox}{max width=0.96\textwidth}
\renewcommand{\arraystretch}{1}
\begin{tabular}{l|ccccc|ccccc}
\hline
 & \multicolumn{5}{c|}{\textbf{Native}} & \multicolumn{5}{c}{\textbf{Projected from English}} \\ 
 & \textbf{Val.} & \textbf{Train} & \textbf{Test} & \textbf{\#Docs} & \textbf{\#Tokens} & \textbf{Val.} & \textbf{Train} & \textbf{Test} & \textbf{\#Docs} & \textbf{\#Tokens}\\ \hline
Spanish & 5 & 36 & 40  & 81 & 24,681 & --- & --- & ---  & --- & ---  \\
French & 5 & 36 & 40  & 81 & 25,196 & --- & --- & ---  & --- & ---  \\
Basque & 5 & 34 & 51  & 90 & 22,505 & --- & --- & ---  &  --- & ---  \\
English & 5 & 37 & 42 & 84 & 29,359 & --- & --- & ---  & --- & --- \\
Italian & 5 & 36 & 45 & 86 & 24,319 & 5 & 37 & 42  & 84 & 32,381 \\
Greek & --- & --- & ---  & ---  & ---  & 5 & 37 & 42  & 84 & 30,069\\
Polish & --- & --- & --- & ---  & ---  & 5 & 37 & 42  & 84 & 27,881\\
Slovak & --- & --- & --- & ---  & ---  & 5 & 37 & 42  &  84 & 26,658\\
Slovenian & --- & --- & --- & ---  & ---  & 5 & 37 & 42  & 84 & 28998\\ \hline
\end{tabular}
\end{adjustbox}
\end{table*}


 
\subsection{E3C tasks}
\label{tasks}

\begin{figure}
    \centering
    \includegraphics[scale=0.33]{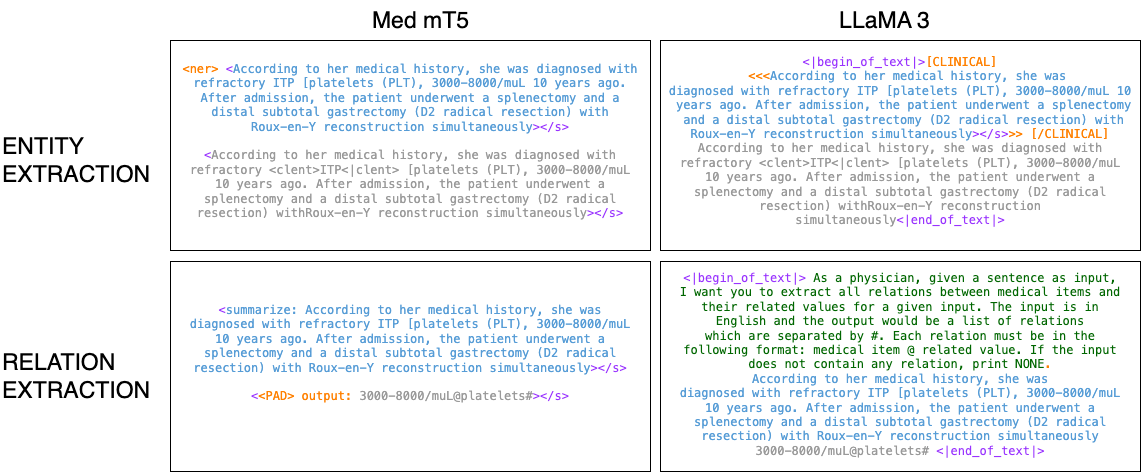}
    \caption{Example of training sequences on Entity Detection and Relation Extraction for MedMT5 and Lllama 3. In \textcolor{violet}{violet} the special tokens required by the models due to their training procedures; in \textcolor{blue}{blue} the input sentence; in \textcolor{orange}{orange} our custom tokens implemented to direct the generation; in \textcolor{gray}{gray} the output formatted following the structure proposed by previous works (Section \ref{tasks}).
    For Relation Extraction using Llama, preliminary experiments indicated that incorporating a system prompt, such as the one highlighted in \textcolor{ForestGreen}{green}, enhances performance. As a result, we adopted this approach. For Entity Extraction on the same model, the system prompt has been substituted by the special token \textit{[CLINICAL]}. }
    \label{fig:prompt-struct}
\end{figure}

With respect to the rich set of E3C annotations depicted in Section \ref{e3c-annotations}, we selected two subtasks which are commonly addressed by the community: 


\begin{itemize}
    \item \textbf{Task 1 - Clinical Entity Detection: }
This task consists in identifying relevant clinical entities in clinical cases annotated as presented in Figure \ref{fig:e3cAnnotations}. We frame the Clinical Entity Detection task as a text-to-text generation task, emphasizing the identification and labeling of textual spans as named entities within a context. Therefore, the sentence presented in as training sequence to the model in the form of a tagged string as presented in the top boxes of Figure \ref{fig:prompt-struct}.
\item \textbf{Task 2 - Test-Result Relation Extraction: }
This task consists in identifying the relations between laboratory tests and their measurements within a clinical note. 
\end{itemize}

 

\subsection{Models} \label{subsect:model}
We conduct our experiments using one representative model from the encoder-decoder family and another from the decoder-only family, respectively. In addition, we focus on models of relatively small size, as fine-tuning and inference are manageable with limited computational infrastructure, often available in industry and academy. For these reasons, we selected MedMT5 (an encoder-decoder model based on mT5), and Llama3-8B. 

\begin{itemize}
    \item \textbf{MedMT5}
\citep{garcia-ferrero-etal-2024-medmt5-open} is the first open-source text-to-text multilingual model explicitly tailored for the medical domain. It is developed by continuously pretraining mT5 \cite{xue-etal-2021-mt5} checkpoints on a specialized medical corpus containing 3 billion words across four languages: English, Spanish, French, and Italian. This domain-specific adaptation improves the model’s ability to handle clinical tasks effectively, leveraging its encoder-decoder architecture. The model is released in two sizes: 	Medical-mT5-large (738M) and Medical-mT5-xl (3B). Recent studies~\citep{bioNLP2024} have demonstrated that MedMT5 achieves promising results in clinical information extraction tasks, outperforming comparable models with alternative pretraining objectives such as traditional fine-tuning, continuous pretraining in the medical domain, and instruction-tuning. 

\item \textbf{Llama 3}
\cite{DBLP:journals/corr/abs-2407-21783} is a family of open-weight decoder-only foundation models pretrained on 15 trillion tokens from multilingual data. The models are released\footnote{\url{https://huggingface.co/meta-llama}} in five sizes in both the base and the instruction-tuned versions. Namely, the base models are 8B and 70B, 405B (version 3.1), 1B and 3B (version 3.2). Llama 3 incorporates a few optimizations, including a 128K token vocabulary and the adoption of grouped query attention.
We used the 8B version for our experiments, as this size is often referred to as very common for benchmarking techniques and models on different datasets due to the relative ease of usage on small GPUs and good performance \cite{wang2024mmluprorobustchallengingmultitask}.
\end{itemize}


\subsection{Fine-tuning procedure} \label{subsect:fine_tuning_procedure}
LLM fine-tuning has been performed using the HuggingFace libraries (transformers\footnote{\url{https://huggingface.co/docs/transformers/index}}, peft\footnote{\url{https://huggingface.co/docs/peft/index}}, and trl\footnote{\url{https://huggingface.co/docs/trl/index}})
on a single A40 (46GB). On MedMT5, we performed fine-tuning on all the model's parameters, while for Llama we used LoRA \cite{hu2021loralowrankadaptationlarge} targeting all the linear modules to reduce the computational costs.

\begin{itemize}
    \item \textbf{MedMT5 for Entity Detection: }
Fine-tuning of the MedMT5 model was carried out using a batch size of 4, a maximum token length of 128 for both input and output, a dropout rate of 0.05, a warmup ratio of 0.06, and an Adam optimizer with an epsilon of 1e-8. We experimented with various learning rates, including 1e-4, 2e-4, 3e-4, 2e-5, and 3e-5, ultimately identifying 1e-4 as the optimal choice. The fine-tuning process spanned up to 25 epochs, with early stopping applied based on validation loss to ensure overfitting was avoided. 

\item \textbf{Llama 3 for Entity Detection: }
Fine-tuning of Llama 3 8B has been performed for $10$ epochs with early stopping on the validation loss, batch size of $16$, using paged Adamw quantized in 8 bit as opitimizer \cite{loshchilov2017decoupled}, and a constant learning rate of $2e-4$. We selected the best model on the development set over $18$ different configurations of LoRA parameters, varying the rank ($16$, $32$, and $64$), alpha ($16$, $32$, and $64$), and the dropout ($0.01$, $0.05$).

\item \textbf{MedMT5 for Relation Extraction: }
Fine-tuning of the MedMT5 model was carried out using a batch size of 2 with gradient accumulation step 2, a maximum token length of 400, a dropout rate of 0.05, a warmup step 1000, an AdamW optimizer with an epsilon of 1e-8 and learning rate  1e-4. The fine-tuning process spanned up to 100 epochs, with early stopping applied based on validation loss to ensure overfitting was avoided. 

\item \textbf{Llama 3 for Relation Extraction: }
LoRA (128*64) Fine-tuning of Llama 3 8B has been performed for 100 epochs with early stopping on the validation loss, the global batch size of 8 and micro batch size of 4, using  AdamW  , a constant learning rate of 3e-4, drop out of 0.05 , the rank of 64 and alpha of 128.

\end{itemize}

\section{Experiments: Unrevised vs Revised Data}
\label{sec:experiments_revised}

In this section, we present experiments on data in projected languages, comparing model performance when trained on the data before and after manual revision, and tested on the revised version. 
The goal here is to quantify the improvement due to the revision process. 

\begin{table*}[!ht]
\centering
\caption{\textbf{Clinical Entities Results on Test Set} on Extended Languages. The delta represents the difference in F1 scores across the five target languages, comparing the performance of the revised and unrevised versions of MedMT5 and Llama 3. Bold values indicate the highest scores achieved by each model for each language across the Unrevised and Revised setups.}
\label{tab:e3c_test_entities}
\begin{adjustbox}{max width=0.9\textwidth}
\renewcommand{\arraystretch}{1}
\begin{tabular}{cc|ccc|ccc|c}
\hline
\multicolumn{2}{c|}{\textbf{Dataset →}} & \multicolumn{3}{c|}{\textbf{Training on Unrevised,}} & \multicolumn{3}{c}{\textbf{Training \& Test}} & \multicolumn{1}{|c}{\textbf{Delta}}\\
\textbf{Languages} & \textbf{Models} & \multicolumn{3}{c|}{\textbf{Test on Revised}} & \multicolumn{3}{c}{\textbf{on Revised}} & \multicolumn{1}{|c}{ }\\  
 \textbf{ ↓} & \textbf{ ↓} & \textbf{P} & \textbf{R} & \textbf{F1} & \textbf{P} & \textbf{R} & \textbf{F1} & \textbf{F1}\\ \hline
\textbf{Greek}     & \textit{MedMT5}   & 52.60 & 38.26 & \textbf{44.30} & 52.68 & 37.11 & 43.55 & -0.75 \\
                   & Llama 3           & 48.80 &  55.80  & 52.01  & 51.34  & 54.79 & \textbf{53.03} & +1.02\\ \hline
\textbf{Polish}    & \textit{MedMT5}   & 56.67 & 41.08 & 47.63 & 54.41 & 45.91 & \textbf{49.80} & +2.17 \\
                   & Llama 3           & 49.2  & 60.3  & \textbf{54.2}  & 56.7  & 51.7  & 54.1 & -0.1 \\ \hline
\textbf{Slovak}    & \textit{MedMT5}   & 51.56 & 42.94 & \textbf{46.86} & 56.02 & 34.57 & 42.76 & - 4.1\\
                   & Llama 3           & 55.6  & 40.1  & 46.6  & 46.0  & 60.4  & \textbf{52.2} & + 5.6\\ \hline
\textbf{Slovenian} & \textit{MedMT5}   & 52.63 & 39.03 & 44.82 & 56.65 & 49.07 & \textbf{52.59} & +7.67\\
                   & Llama 3           & 49.2  & 53.8  & \textbf{51.4}  & 42.1  & 62.2  & 50.2 & -1.2\\ \hline
\textbf{Italian}   & \textit{MedMT5}   & 65.97 & 47.57 & 55.28 & 60.55 & 58.05 & \textbf{59.27} & +3.99 \\
                   & Llama 3           & 61.1  & 58.4  & 59.7  & 62.6  & 59.4  & \textbf{61.0}  & +1.3\\ \hline
\textbf{English}   & \textit{MedMT5}   & -& -& -& 57.19 & 62.73 & 59.83 & - \\ 
\textbf{(Source)}  & Llama 3           & - & -& -& 55.8  & 50.7 & 53.2 & -\\ \hline
\end{tabular}
\end{adjustbox}
\end{table*}

\subsection{Clinical Entities Detection}

We evaluated the projected dataset for all languages using MedMT5 and Llama 3, comparing results between versions before (Unrevised) and after manual revisions (Revised), as summarized in Table~\ref{tab:e3c_test_entities}. The findings reveal a dynamic interplay between domain-specific specialization in MedMT5 and the general-purpose adaptability of Llama 3. MedMT5 shows notable improvements between the two dataset versions in languages closer to its pretraining corpus, such as Slovenian (+7.67) and Italian (+3.99). These gains highlight the effectiveness of manual revisions in addressing projection errors, such as token misalignments and boundary inconsistencies, thereby boosting precision and recall. Conversely, Llama 3 exhibits smaller deltas in languages like Polish and Greek, reflecting its robustness and broader adaptability across diverse linguistic contexts, albeit with lower sensitivity to domain-specific corrections. Its performance in Slovak (+5.6) further underscores its capacity to leverage revisions effectively, even in languages outside the core medical domain.

However, both models encounter challenges in certain cases. MedMT5 shows performance drops in Greek (-0.75) and Slovak (-4.1), possibly due to overfitting to revised patterns or mismatches with test set distributions. Similarly, Llama 3 records a decline in Slovenian (-1.2), suggesting that revisions might inadvertently introduce patterns that are less representative of the broader test set, leading to performance declines. Overall, MedMT5’s reliance on high-quality, domain-specific annotations leads to significant performance shifts, while Llama 3’s general-purpose pretraining ensures steadier outcomes across a broader linguistic spectrum. These results emphasize the critical need to balance domain-specific adaptation with multilingual generalization to optimize model performance in cross-lingual clinical tasks.

\begin{table*}[!ht]
\centering
\caption{\textbf{Relation Extraction Results on Test Set} on Extended Languages. The delta represents the difference in F1 scores across the five target languages, comparing the performance of the revised and unrevised versions of MedMT5 and Llama 3. Bold values indicate the highest scores achieved by each model for each language across the Unrevised and Revised setups.}
\label{tab:e3c_test_relations_Un_re}
\begin{adjustbox}{max width=0.9\textwidth}
\renewcommand{\arraystretch}{1}
\begin{tabular}{cc|ccc|ccc|c}
\hline
\multicolumn{2}{c|}{\textbf{Dataset →}} & \multicolumn{3}{c|}{\textbf{Training on Unrevised,}} & \multicolumn{3}{c}{\textbf{Training \& Test}} & \multicolumn{1}{|c}{\textbf{Delta}}\\
\textbf{Languages} & \textbf{Models} & \multicolumn{3}{c|}{\textbf{Test on Revised}} & \multicolumn{3}{c}{\textbf{on Revised}} & \multicolumn{1}{|c}{ }\\  
 \textbf{ ↓} & \textbf{ ↓} & \textbf{P} & \textbf{R} & \textbf{F1} & \textbf{P} & \textbf{R} & \textbf{F1} & \textbf{F1}\\ \hline
\textbf{Greek}     & \textit{MedMT5}   & 48.76& 35.22& \textbf{40.90}& 45.12& 33.13& 38.20& -2.7\\
                   & Llama 3           & 50.90& 33.43& 40.36& 57.62& 40.59& \textbf{47.63}& +7.27\\ \hline
\textbf{Polish}    & \textit{MedMT5}   & 47.91& 27.71& 35.11& 41.04& 40.40& 40.54& +5.43\\
                   & Llama 3           & 54.31& 38.29& \textbf{44.91}& 56.66& 40.96& \textbf{47.55}& +3.04\\ \hline
\textbf{Slovak}    & \textit{MedMT5}   & 43.16& 31.79& 37.05& 43.61& 32.08& 37.15& +0.1 \\
                   & Llama 3           & 60.68& 42.38& \textbf{49.91}& 55.05& 47.16& \textbf{50.80}& +0.89\\ \hline
\textbf{Slovenian} & \textit{MedMT5}   & 46.93& 34.53& 39.79& 48.42& 41.44& 44.66& +5.47\\
                   & Llama 3           & 44.94& 39.20& \textbf{46.27}& 53.52& 50.19& \textbf{51.78}& +5.51\\ \hline
\textbf{Italian}   & \textit{MedMT5}   & 41.96& 31.94& 36.27& 35.19& 48.16& 41.96& +5.69\\
                   & Llama 3           & 56.26& 36.41& \textbf{42.57}& 61.97& 48.65& \textbf{54.51}& +11.94\\ \hline
\textbf{English}   & \textit{MedMT5}   & - & - & - & 59.19& 36.22& 45.00& - \\ 
\textbf{(Source)}  & Llama 3           & - & - & - & 68.28& 47.98& 56.36& - \\ \hline
\end{tabular}
\end{adjustbox}
\end{table*}

\subsection{Relation Extraction}

As reported in Table \ref {tab:e3c_test_relations_Un_re}, for MedMT5, the impact of the revision of the dataset varies between languages. Greek shows a slight decline in F1 (-2.7), with precision decreasing by -3.64 and recall by -2.09, indicating potential challenges during porting or overfitting. Polish sees a notable F1 improvement (+5.43), driven by a substantial recall increase (+12.69), likely attributed to better handling of boundary cases or underrepresented relations in the unrevised dataset that were clarified in the revised version. Slovenian achieves a solid F1 gain (+5.47), largely due to improved precision by +1.49 and recall by +6.91. Italian shows the most significant improvement (+5.69), with precision decreasing by -6.77 but recall improving dramatically by +16.22. This marked improvement in recall suggests that the revisions likely addressed errors in boundary cases or data sparsity issues, enabling the model to better capture a wider range of correct relations even at the cost of slight precision loss. Slovak exhibits only a marginal increase in F1 (+0.1), where a precision gain of +0.45 is balanced by a slight recall increase of +0.29, suggesting a minimal impact from overall revisions of the dataset. This variability reflects language-specific dynamics in translation quality, model adaptation, and the effectiveness of fine-tuning.
Llama3 8B demonstrates consistent improvements in all languages after revision, with Italian seeing the largest F1 boost (+11.94), driven by substantial enhancements in both precision (P: +5.71) and recall (R: +12.24). Greek improves significantly (+7.27), with precision increasing by +6.72 and recall by +7.16, showcasing robust fine-tuning capability with revised data. Slovenian and Polish also exhibit notable gains (+5.51 and +3.04, respectively). Slovenian benefits from a precision improvement of +8.58 and a recall increase of +11.99, while Polish gains more limitedly in both metrics (P: +2.35, R: +2.67). The Slovak marginal increase (+0.89) reflects a balance between slightly reduced precision (-5.63) and improved recall (+4.78), aligning with MedMT5 findings and reinforcing stability across both models despite different behaviors in handling precision and recall trade-offs.
Llama3 outperforms MedMT5 in nearly all metrics, highlighting its larger capacity and advanced architecture. The performance gap is especially pronounced in Italian and Greek, where the ability of Llama3 to leverage revised datasets provides stronger results. Although MedMT5 struggles with Greek and sees modest improvements for Slovak, Llama3 demonstrates consistent adaptability. These results underscore the value of advanced models like Llama3 and the critical role of dataset quality and revisions.


\section{Experiments: Cross-language Transfer Learning}
\label{sec:experiment_all_languages}

In this section we present experiments on data from projected languages, comparing model performance when independently trained on single languages and when jointly trained on all languages.

\subsection{Clinical Entities Detection}

In this experiment, we fine-tuned unified versions of MedMT5 and Llama 3 using a combined dataset that integrates training and validation data across all target languages. This augmented multilingual training setup was then evaluated on language-specific test sets to measure its impact compared to the performance on the revised datasets. By training on a richer and more diverse corpus, the goal was to assess how exposure to multilingual data influences model generalization and performance in clinical entity recognition tasks. The results of this evaluation are presented in Table~\ref{tab:clinical_entities_augmented_test}.

\begin{table*}[ht!]
\centering
\caption{\textbf{Clinical Entities Results on Test Set} with Augmented Training. The delta represents the difference in F1 scores across the languages, comparing the performance of the revised (\textit{Table~\ref{tab:e3c_test_entities}}) and augmented setup (\textit{this table}) of MedMT5 and Llama 3.}
\label{tab:clinical_entities_augmented_test}
\begin{adjustbox}{max width=0.9\textwidth}
\renewcommand{\arraystretch}{1}
\setlength\tabcolsep{9pt}
\begin{tabular}{c|cccc|cccc}
\hline
\multicolumn{9}{c}{\textbf{Train \& Validation on Augmented Train and Augmented Development Sets, }} \\
\multicolumn{9}{c}{\textbf{respectively. Evaluation on independent Test sets}} \\ \hline
\multicolumn{1}{c|}{\textbf{Models →}} & \multicolumn{4}{c|}{\textbf{MedMT5}} & \multicolumn{4}{c}{\textbf{Llama 3}} \\
\textbf{Languages ↓} & \textbf{P} & \textbf{R} & \textbf{F1} & \textbf{Delta} & \textbf{P} & \textbf{R} & \textbf{F1} & \textbf{Delta}\\ 
\hline
\textit{\textbf{Greek}} & 60.1 & 41.5 & 49.1 & +5.6 & 57.2 & 55.3 & 56.2 & +3.2\\
\textit{\textbf{Polish}} & 58.9 & 49.3 & 53.6 & +3.8 & 64.7  & 48.9 & 55.7 & +1.6\\
\textit{\textbf{Slovak}} & 56.5 & 51.1 & 53.7 & +10.9 & 58.2 & 52.1 & 55.0 & +2.8\\
\textit{\textbf{Slovenian}} & 58.6 & 53.4 & 55.8 & +3.3 & 59.5 & 47.1 & 52.6 & +2.4\\
\textit{\textbf{Italian}} & 60.3 & 66.1 & 63.1 & +3.8 & 66.2 & 57.6 & 61.6 & +0.6\\
\textit{\textbf{English (Source)}} & 62.7 & 65.2 & 63.9 & +4.1 & 65.5 & 54.5 & 59.5 & +6.3 \\ \hline
\textit{\textbf{Average}} & 59.5 & 54.4 & 56.6 & +5.2 & 61.6 & 52.1 & 56.3 & +2.2\\
\hline
\end{tabular}
\end{adjustbox}
\end{table*}

Both MedMT5 and Llama 3 show notable improvements with this approach, demonstrating the efficacy of augmented training. MedMT5 achieves significant F1 score gains, particularly in Slovak (+10.9) and Greek (+5.6), reflecting its ability to leverage domain-specific data from multiple languages to address prior limitations such as overfitting and token misalignment. Languages closer to its pretraining corpus, such as Italian (+3.8) and Slovenian (+3.3), also benefit, emphasizing MedMT5's strength in domain-specific scenarios when supplemented with diverse data. Llama 3, on the other hand, demonstrates robust adaptability across all languages, with notable improvements in English (+6.3), Greek (+3.2), and Slovak (+2.8). While its F1 score deltas are generally smaller than MedMT5's, Llama 3 exhibits balanced gains across languages, indicating its resilience to data variability and its general-purpose training's ability to adapt effectively. These results underscore the complementary strengths of both models: MedMT5 excels in leveraging domain-specific nuances from augmented data, while Llama 3 showcases steady improvements across linguistic diversity.

\subsection{Relation Extraction}

\begin{table*}[ht!]
\centering
\caption{\textbf{Relation Extraction Results on Test Set} with Augmented Training. The delta represents the difference in F1 scores across the languages, comparing the performance of the revised (\textit{Table \ref {tab:e3c_test_relations_Un_re}}) and augmented setup (\textit{this table}) of MedMT5 and Llama 3.}
\label{tab:relation_extraction_medmt5_llama_test}
\begin{adjustbox}{max width=0.9\textwidth}
\renewcommand{\arraystretch}{1}
\setlength\tabcolsep{8pt}
\begin{tabular}{c|cccc|cccc}
\hline
\multicolumn{9}{c}{\textbf{Train \& Validation on Augmented Train and Augmented Development Sets, }} \\
\multicolumn{9}{c}{\textbf{respectively. Evaluation on independent Test sets}} \\ \hline
\multicolumn{1}{c|}{\textbf{Models →}} & \multicolumn{4}{c|}{\textbf{MedMT5}} & \multicolumn{4}{c}{\textbf{Llama 3}} \\
\textbf{Languages ↓} & \textbf{P} & \textbf{R} & \textbf{F1} & \textbf{Delta} & \textbf{P} & \textbf{R} & \textbf{F1} & \textbf{Delta}\\ 
\hline
\textit{\textbf{Greek}} & 47.88 & 40.59 & 43.94 & +5.74 & 61.81      & 40.59      & 49.00 & +1.37\\
\textit{\textbf{Polish}} & 42.96 & 35.85 & 39.08 & -1.54  & 57.20      & 41.86      & 48.34 & +0.79\\
\textit{\textbf{Slovak}} & 57.14 & 44.17 & 49.83 & +12.68  & 60.86      & 41.72      & 49.55 & -1.25\\
\textit{\textbf{Slovenian}} & 51.34 & 40.24 & 45.11 & +0.45  & 61.00      & 44.90      & 51.92 & +0.14\\
\textit{\textbf{Italian}} & 48.56 & 40.29 & 44.04 & +2.08 & 58.49      & 45.48      & 51.09 & -3.42\\
\textit{\textbf{English (Source)}} & 49.23 & 39.93 & 44.10 & -0.90 &54.83       & 42.10      & 47.63  & -8.73 \\ \hline
\textit{\textbf{Average}} & 49.51& 40.17& 44.35& +3.23& 59.03& 42.77& 49.58& -1.85\\
\hline
\end{tabular}
\end{adjustbox}
\end{table*}

From Table \ref{tab:relation_extraction_medmt5_llama_test}, the results of MedMT5 fine-tuning reveal varying degrees of improvement across languages when switching from individual fine-tuning to fine-tuning on the augmented ``all languages" dataset. In particular, Slovak shows the largest gain in F1 (+12.68), driven by substantial increases in precision (+13.53) and recall (+12.09), highlighting the strong benefit of cross-linguistic knowledge transfer in addressing language-specific challenges. Greek also experiences a significant increase in F1 (+5.74), with precision and recall improving by +2.76 and +7.46, respectively. While Italian sees a moderate improvement (+2.08), its increase in precision (+13.37) is offset by a decline in recall (-7.87), reflecting a precision-focused adjustment. Polish and English exhibit slight declines in F1 (-1.54 and -0.90), where modest precision gains (+1.92 and -9.96) are overshadowed by decreases in recall (-4.55 and +3.71). Slovenian achieves a negligible F1 improvement (+0.45), with a small precision gain (+2.92) counterbalanced by a slight decline in recall (-1.20). These findings suggest that MedMT5 benefits greatly from multilingual fine-tuning, especially for lower-resource languages like Slovak.

Llama3's performance shows consistent F1 improvements in Greek (+1.37) and Polish (+0.79) when fine-tuned on the ``all languages" dataset, benefiting from small gains in both precision and recall. In contrast, Italian and English experience notable declines (-3.42 and -8.73 F1, respectively), with both languages showing drops in precision and recall. Italian recall falls significantly (-3.17), and English faces a substantial decline in recall (-5.88), potentially due to negative knowledge transfer effects. For Slovak, the switch results in a slight decrease in F1 (-1.25) despite an increase in precision (+5.81), as recall suffers (-5.44). Slovenian, however, experiences minimal overall change (+0.14 F1), with a precision gain of +7.48, balancing a recall drop of -5.29. These results underscore the sensitivity of Llama3's performance to negative transfer effects in some languages while still harnessing cross-linguistic benefits for others like Greek and Polish.
While both models benefit from multilingual fine-tuning, their performance dynamics differ. MedMT5 demonstrates broader improvements, with Slovak, Greek, and Italian reaping significant F1 gains. Llama3, on the other hand, exhibits more sensitivity to language-specific effects, showing notable declines for Italian and English but steady improvement for others. These observations highlight the trade-offs in precision and recall optimization, as well as the differing strengths of each architecture in leveraging cross-linguistic transfer effectively.

\section{Experiments: Native vs Projected Data}
\label{sec:italian_native_vs_projected}

We aimed to investigate whether the information embedded in projected data could meaningfully support inference on real-world data, rather than being confined to synthetic scenarios. To explore this, we focused on Italian data, leveraging the availability of both native and projected datasets. Our objective was to assess whether models trained on projected data could perform on par with those trained on authentic data when tested in real-world settings. We run these experiments using Llama, as it showed to achieve the best performances in Section \ref{sec:experiments_revised}. We compare results achieved by Llama fine-tuned following the procedure described in Section \ref{subsect:fine_tuning_procedure} on native Italian data, projected Italian data, a combination of the two, and on all the projected data.

\begin{table}[ht!]
\centering
\caption{Clinical Entity Detection and Relation Extraction on native and projected data for Italian. The test is performed only on Italian, while the training includes only Italian (first two rows) or all the projected languages (all lang.).}
\label{tab:combined_performance}
\begin{tabular}{c|p{4cm}|c|c|c}
\hline
&  & \multicolumn{3}{c}{\textbf{Test (Italian)}} \\ \cmidrule(lr){3-5}
\textbf{}                              &    & \textbf{Native} & \textbf{Projected} & \textbf{Average} \\ 

\hline
 \hline 
& \textbf{Clinical Entities Detection} & F1 & F1 & F1\\ \hline
\multirow{3}{*}{\rotatebox[origin=c]{90}{\textbf{Train}}} 
 & \textbf{Native (Italian)} & 64.5 & 55.3 & 59.9 \\
 & \textbf{Projected (Italian)} & 56.7 &59.2 & 58.0 \\
 & \textbf{Projected (all lang.)}  &  62.5  & 61.6 & 62.1 \\ \hline
 \hline 
\\
 \hline 
& \textbf{Relation Extraction} & F1 & F1 & F1\\ \hline
\multirow{3}{*}{\rotatebox[origin=c]{90}{\textbf{Train}}} 
 & \textbf{Native (Italian)} &  71.68 &  46.73 & 59.2 \\
 & \textbf{Projected (Italian)} & 46.28 &  54.5 & 50.4 \\
 & \textbf{Projected (all lang.)}  & 44.14 & 51.1 & 47.6 \\ \hline
\end{tabular}
\end{table}

Results for Clinical Entity Detection and Relation Extraction are shown in Table \ref{tab:combined_performance}.
The primary observation from the results on the Clinical Entity Detection task is that the model trained on projected data consistently underperformed compared to the one trained on native data when evaluated on native datasets. This discrepancy is likely attributable to the closer alignment in distribution between the training and testing native data. The performance gap, quantified as a delta of $7.8$ points, highlights that, while projected data can be useful, it is not as effective as native data when the latter is available for training. Interestingly, the model trained on augmented data, combining all the projected languages, significantly narrowed the performance gap, reducing it from $7.8$ to $2.0$ points. This finding suggests that when projected data is effectively augmented, it can provide sufficient information for a model to achieve performance levels comparable to those trained on native data.

Results for the relation extraction task demonstrate that models trained on native Italian data achieve the highest performance on the native test set (F1 = 71.68) but perform poorly on projected data (F1 = 46.73), revealing limited generalization to different data distributions. Conversely, models trained on projected Italian data perform better on projected datasets (F1 = 54.5) but struggle with native test data (F1 = 46.28). 

\section{Conclusion}
We have presented a multilingual dataset in the medical domain, consisting of clinical cases annotated with diseases and test-result relations. The dataset includes both native texts in five languages (English, French, Italian, Spanish and Basque) and texts translated and projected from the English source into five target languages (Greek, Italian, Polish, Slovak, and Slovenian). A two-step semi-automatic approach has been implemented: automatic annotation projection based on LLMs followed by human revision. We have performed several experiments showing that current state-of-the-art LLMs can benefit from being fine-tuned on the dataset. We have also shown that transfer learning in different languages is very effective, mitigating the scarcity of data. Finally, we have compared performance both on native data and on projected data.

\section*{Acknowledgments}
This work has been partially funded by the European Union under the Horizon Europe eCREAM Project (Grant Agreement No.101057726) and IDEA4RC Project (Grant Agreement No.101057048). Views and opinions expressed are however those of the authors only and do not necessarily reflect those of the European Union or the European Health and Digital Executive Agency (HADEA). Neither the European Union nor the granting authority can be held responsible for them.
This work was supported by the CHISTERA grant of the Call XAI 2019 of the ANR with the grant number Project-ANR-21-CHR4-0002.
We acknowledge the support of the PNRR project FAIR - Future AI Research (PE00000013), under the NRRP MUR program funded by  NextGenerationEU.



\begin{appendices}




\end{appendices}


\bibliography{main}

\end{document}